\documentclass[letterpaper, 10 pt, conference]{ieeeconf} 

\IEEEoverridecommandlockouts 

\usepackage{graphics} 
\usepackage{epsfig} 
\usepackage{mathptmx} 
\usepackage{times} 
\usepackage{amsmath} 
\usepackage{amssymb} 
\usepackage[utf8]{inputenc}
\usepackage[hyphens]{url}
\usepackage{multirow}
\usepackage{booktabs} 

\title{\LARGE \bf Distortion-Aware PETR for BEV Object Detection with Mixed
Pinhole–Fisheye Cameras }

\author{Xiangzhong Liu
\thanks{Xiangzhong Liu is with Machine Learning Group, fortiss GmbH,
Guerickestraße 25, 80805 Munich, Germany {\tt\small xiangzhong.liu@tum.de}}%
}
\usepackage{textcomp}
\usepackage[absolute,showboxes]{textpos}
\TPMargin{4pt}
\newcommand{\copyrightstatement}{
    \begin{textblock}{0.84}(0.08,0.938)
         \noindent{\footnotesize{\copyright 2026 IEEE.
         Personal use of this material is permitted.
         Permission from IEEE must be obtained for all other uses, in any current or future media, including reprinting/republishing this material for advertising or promotional purposes, creating new collective works, for resale or redistribution to servers or lists, or reuse of any copyrighted component of this work in other works.

         \noindent
         Accepted for publication in Proceedings of the IEEE International Conference on Robotics and Automation (ICRA),  Vienna, Austria, 1-5 June 2026.}}
    \end{textblock}
}  

\begin{document}
    \maketitle
    \copyrightstatement
    \thispagestyle{empty}
    \pagestyle{empty}

    \begin{abstract}
        Fisheye cameras are widely deployed in autonomous driving perception suites
        for their low cost and full-coverage field of view (FOV), yet their
        potential remains under-leveraged in 3D object detection. Severe radial distortion
        challenges most BEV detectors by violating the fundamental assumption of
        uniform sampling. To bridge this gap, we propose Distortion-Aware PETR (DAPETR),
        a projection-free detector tailored for mixed pinhole–fisheye camera
        setups. DAPETR incorporates two key learned-adaptive modules: a unified
        distortion-aware positional embedding that harmonizes positional
        encodings for image representations with fisheye geometry, and a bidirectional
        feature-geometry co-modulation module that mutually adapts image features
        and 3D positional embeddings. In our experiments on a converted KITTI-360
        benchmark, we systematically compare our learned-adaptive approach against
        PETR in polar coordinates (PolarPETR). We find that while both methods
        improve over the baseline, our learned modules achieve superior performance.
        Crucially, we uncover a negative interaction when combining both strategies,
        revealing that learned adaptation and explicit geometric re-parameterization
        can conflict. Our final DAPETR model significantly advances the research
        and benchmark for fisheye BEV detection, providing critical insights
        into effective distortion-aware 3D perception design other than image
        rectification.
    \end{abstract}

    \section{INTRODUCTION}

    Modern autonomous driving systems increasingly rely on mixed camera
    configurations with pinhole and fisheye cameras with ultra-wide FOV coverage
    to achieve full $360^{\circ}$ perception at manageable cost~\cite{kumar2023surround}.
    Despite the wide deployment of fisheye cameras, their potential for 3D
    object detection remains largely under-exploited. The fundamental challenge lies
    in the severe radial distortion that violates the uniform sampling assumption
    underlying most BEV-based detection frameworks~\cite{li2022bevsurvey}.
    Benchmarks like nuScenes~\cite{caesar2020nuscenes} have substantially
    advanced BEV object detection with standard pinhole setups. However, there
    is still no widely adopted real-world benchmark for BEV detection that
    incorporates fisheye cameras.
    \begin{figure}[htbp]
        \centering
        \includegraphics[width=1.0\linewidth]{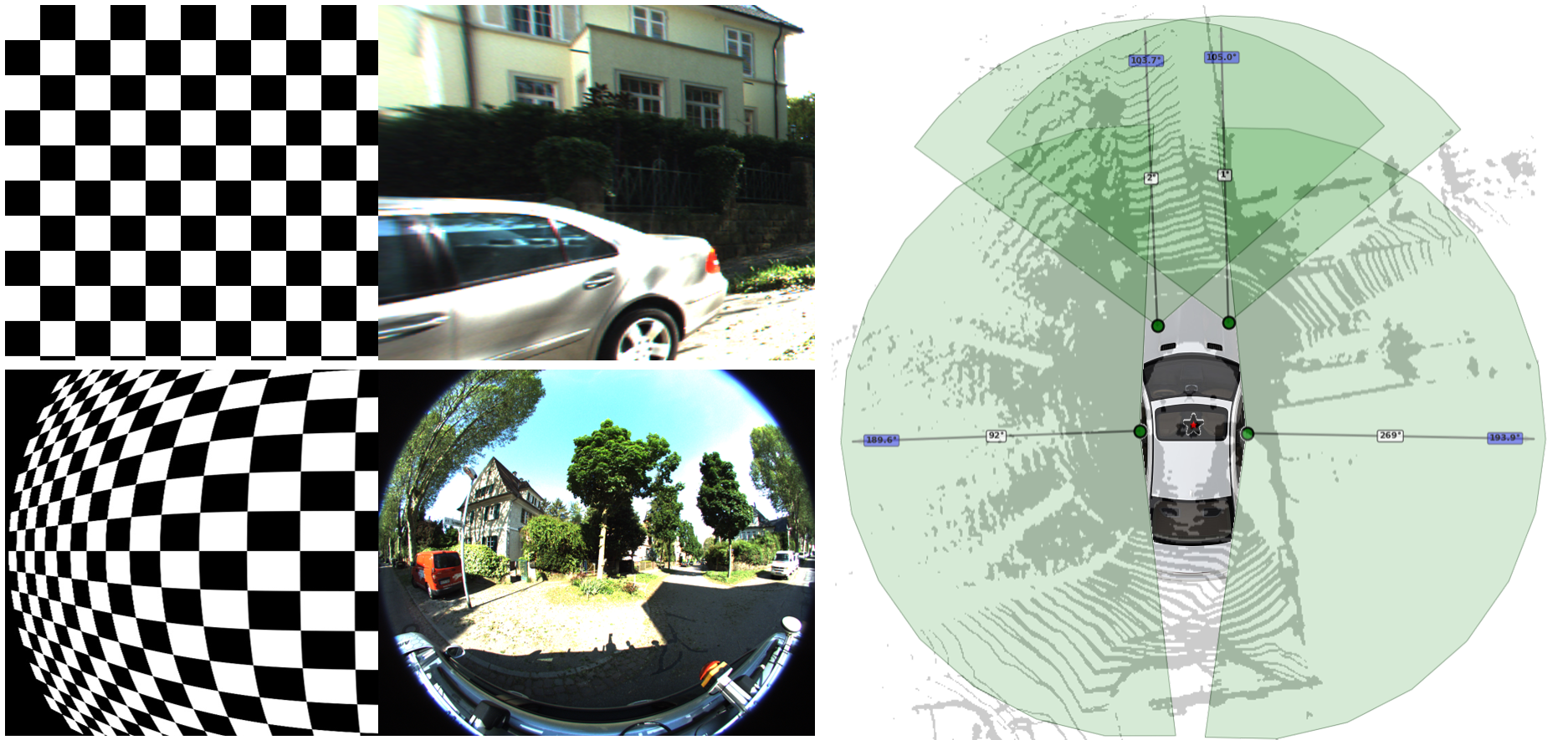}
        \caption{Sensor configuration and full FOV coverage. Left: KITTI-360
        provides both pinhole and fisheye cameras. The distortion 
        characteristics are visualized by their respective checkerboard patterns(uniform grid spacing v.s. severe radial 
        warping ). Right: Top-down view of the KITTI-360 
        sensor setup with 2 front-seeing pinhole cameras and
        2 side fisheye cameras overlapping with roof LiDAR, providing 
        complete $360^{\circ}$ surround coverage.}
        \label{fig:FOV_kitti}
    \end{figure}

    Projection-free methods like PETR~\cite{liu2022petr} bypass explicit
    geometric transformations and instead employ learned spatial priors encoded
    via 3D position embeddings. Implicit BEV approaches become difficult to implement
    facing complex sensor calibration for fisheye cameras. When camera
    parameters describe severe distortion, the fundamental correspondence between
    2D image features and 3D spatial locations deteriorates. This calibration
    sensitivity and geometric breakdown lead to substantial performance
    degradation, preventing projection-free architectures from capitalizing on their
    computational efficiency advantages in mixed-camera systems.

    Polar BEV representations have shown promise for better alignment with camera
    frustum geometry~\cite{jiang2023polarformer, yu2024polarbevdet,chen2025polardetr},
    and naturally match the non-uniform distribution of scene content. However,
    their application has been restricted to improve BEV performance for unified
    pinhole cameras. All three methods employ explicit projection-based view transformation
    modules (VTMs), yet no work has explored adapting projection-free detectors
    to polar space. Specifically, polar representation aligns with fisheye
    geometry by preserving angular consistency across camera frustums and
    enforcing a more uniform sampling density in radial and angular dimensions. Combining
    projection-free efficiency with this polar geometric alignment could simultaneously
    achieve distortion robustness and computational efficiency, yet remains a
    critical unexplored research direction for fisheye 3D perception.

    In this work, we introduce Distortion-Aware PETR (DAPETR), a projection-free
    BEV detector specifically designed for mixed pinhole–fisheye camera setups. We
    employ unified distortion-aware positional embeddings for both 2D pixel tokens
    and 3D position-aware features with a unified camera model. In addition, a bidirectional
    feature–geometry co-modulation module allows distortion-aware image features
    and 3D positional embeddings to refine one another for cross-attention stages.
    As an insightful comparison, we reformulate PETR to operate in polar coordinates
    as PolarPETR. We evaluate our method on the KITTI-360 dataset~\cite{liao2022kitti},
    which provides an ideal testbed with a combination of 2 front pinhole and 2 side
    fisheye cameras.

    Our main contributions are:
    \begin{itemize}
        \item We develop unified distortion-aware positional embeddings for both
            2D pixel tokens and 3D position-aware features through the MEI~\cite{mei2007single}
            camera model, harmonizing image representations with fisheye
            geometry for robust cross-attention under severe distortion.

        \item We introduce bidirectional feature-geometry co-modulation that mutually
            adapts distortion-aware image representations and 3D positional embeddings,
            enabling enhanced appearance-geometry alignment through joint
            spatial reasoning.

        \item We conduct the first systematic comparison between explicit geometric
            alignment (PolarPETR) and learned feature adaptation, revealing a
            negative interaction that provides critical insights for future
            distortion-aware model design.
    \end{itemize}

    \section{RELATED WORK}
    \label{sec:related_works}

    \subsection{BEV Multi-View 3D Detection}
    The BEV representation has become the de facto standard for multi-view 3D
    object detection, providing a unified space for sensor fusion and temporal modeling~\cite{li2022bevsurvey}.
    Current methods primarily differ in their view transformation module (VTM), which
    maps features from image space to the BEV representations.

    Forward projection methods, pioneered by LSS~\cite{philion2020lift}, explicitly
    predict a depth distribution for each image pixel and lift 2D features into 3D
    space. BEVDet~\cite{huang2021bevdet} and its successors~\cite{huang2022bevdet4d}
    optimized this approach, but its accuracy relies heavily on the quality of the
    intermediate depth prediction, which is intractable for distorted fisheye
    images.

    Backward projection approaches, such as BEVFormer~\cite{li2024bevformer} and
    DETR3D~\cite{wang2022detr3d}, pre-define a set of BEV reference/queries and
    project them back to 2D image planes to sample features with deformable
    attention. It avoids explicit depth prediction but restricts the receptive field
    to local area with significant computational cost.

    Projection-free models like PETR~\cite{liu2022petr} offer an alternative bypassing
    explicit feature projection. PETR enriches 2D image features with 3D
    positional embeddings derived from camera intrinsics and extrinsics. A decoder-only
    transformer then attends to spatial-aware features to directly predict 3D bounding
    boxes. While efficient, PETR's reliance on learned geometric priors makes it
    sensitive to camera configurations and calibrations. Our work is the first
    to address this limitation by adapting PETR to handle severe lens distortion.

    \subsection{Polar Coordinate Representation}
    The standard Cartesian BEV grid is ill-suited to the radial nature of the
    camera projections, leading to sparse and inefficient representations.
    Recent works have demonstrated that polar/cylindrical coordinates better
    match the non-uniform distribution of scene content and exploit view
    symmetry in surround-view systems.

    PolarDETR~\cite{chen2025polardetr} explicitly parameterizes 3D objects in
    polar coordinates (radial distance and angle) and decomposes velocities into
    radial/tangential components, achieving faster convergence and better accuracy
    than DETR3D. PolarBEVDet~\cite{yu2024polarbevdet} adapts the BEVDet4D
    framework by replacing the standard Cartesian BEV with polar grids using
    angular-radial bins. It reformulates all core components, including view
    transformation, temporal fusion, and detection head, demonstrating that polar
    BEV representation better aligns with image information density (dense near
    vs. sparse far). PolarFormer~\cite{jiang2023polarformer} similarly adapted forward
    projection methods with transformers to polar BEV space for improved efficiency
    and uniformity.

    These successes motivate our investigation into adapting projection-free detectors
    to a polar space handling fisheye distortion, combining geometric alignment
    advantages with computational efficiency of projection-free designs for
    fisheye 3D perception.
    \begin{figure*}[htbp]
        \centering
        \epsfig{file=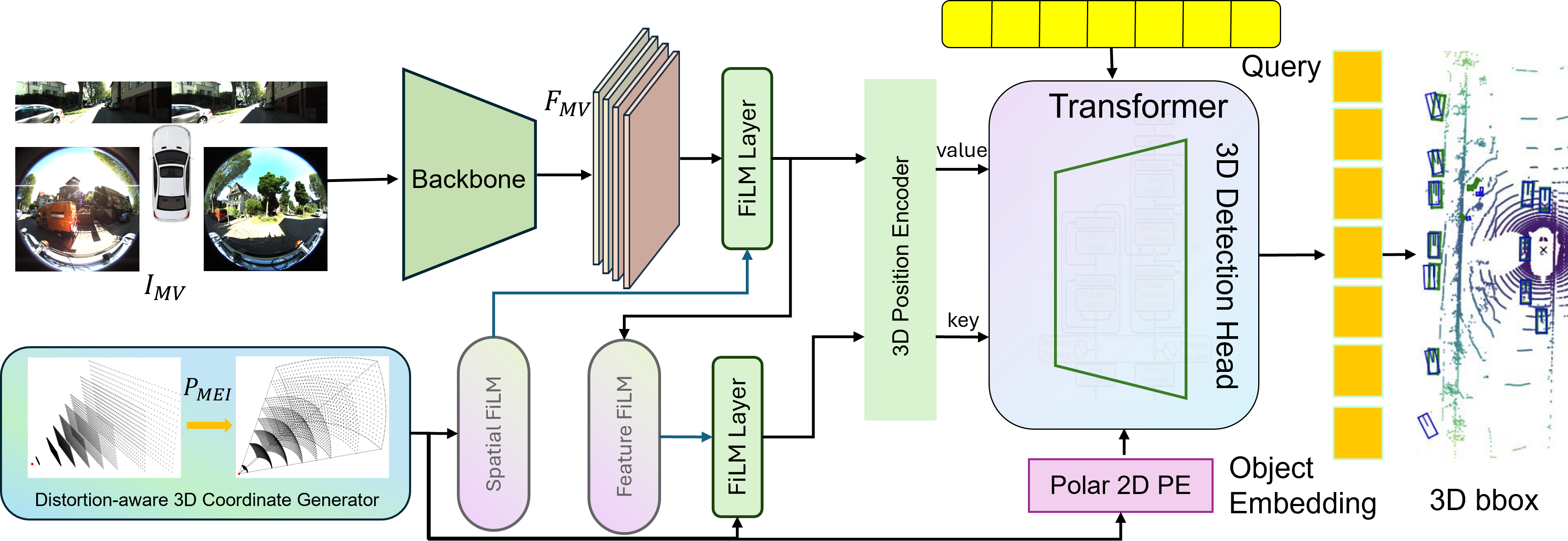,width=0.98\textwidth}
        \caption{An overview of the Distortion-Aware PETR pipeline. Multi-view images
        from mixed pinhole and fisheye cameras are fed into an image backbone.
        MEI camera model is applied for generating 3D coordinates for 3D
        positional encoding. We introduce a polar positional embedding (PE) for the
        2D image features. A co-modulation module (e.g., FiLM) refines the image
        features based on the 3D query positions before the attention mechanism.
        The decoder then outputs 3D bounding box predictions.}
        \label{fig:full_structure}
    \end{figure*}

    \subsection{Distortion-Aware Modeling}
    Handling the severe distortion of fisheye lenses is a long-standing problem
    in computer vision. Traditional methods rely on pinhole or approximate rectification,
    which projects the fisheye image onto a virtual pinhole plane. However, this
    process introduces computational overhead to get projection mappings and
    information loss due to the limited FOV.

    \textbf{Distortion-Aware Architectures:} More recent work has focused on building
    distortion-aware models. F2BEV~\cite{samani2023f2bev} adapted BEVFormer with
    a fisheye 3D-to-2D projection with MEI, but was limited to a backward-projection
    architecture and synthetic fisheye images. Yogamani et al.~\cite{yogamani2024fisheyebevseg}
    studied BEV semantic segmentation from surround-view fisheye cameras,
    showing that simple cylindrical undistortion~\cite{yogamani2019woodscape} is
    suboptimal compared to distortion-aware learnable BEV pooling with occlusion
    reasoning. FisheyeDetNet~\cite{sistu2024fisheyedetnet} addresses 2D fisheye object
    detection through a data-centric approach for performance improvements,
    replacing standard rectangular boxes with rotated polygons for annotations
    in polar coordinates, but does not fundamentally adapt the model architecture
    for fisheye geometry. Beyond that, spherical transformer architectures~\cite{carlsson2024heal}
    extend Vision Transformers to spherical projections, while DarSwin~\cite{athwale2023darswin}
    presents a distortion-aware encoder-only architecture, offering generalized distortion
    modeling.

    \textbf{Camera-Aware Feature Learning:} Another approach incorporates camera
    models directly into networks. Cam-Convs~\cite{facil2019cam} appends per-pixel
    FOV and normalized pixel coordinates to image feature tensors, enabling
    depth networks to generalize across different intrinsics. Reichert et al.~\cite{reichert2024sensor}
    extend this to diverse lenses using the unified camera model by distorting FOV
    maps according to lens parameters to yield distortion-robust features without
    explicit distortion inversion. RectConv~\cite{griffiths2024adapting}
    leverages kernel adjustments such that the convolutional filters see
    rectified patches while retaining the full FOV of fisheye images. Calibrated
    Convolutions~\cite{Berenguel-Baeta_2023_BMVC} proposes deforming convolution
    kernels using the calibration parameters to adapt standard CNNs to fisheye distortion.
    While effective for single-image tasks, per-view adjustments cannot guarantee
    the consistent cross-camera spatial alignment required for multi-view BEV detection
    with heterogeneous sensor configurations.
    \subsection{Feature Modulation for Camera Adaptation}
    Feature-wise modulation techniques such as FiLM~\cite{perez2018film} and
    Squeeze-and-Excitation blocks~\cite{hu2018squeeze} have proven effective for
    adapting visual features to contextual cues in language-guided vision and
    domain adaptation. These modulation networks generate scale and shift parameters
    that rescale intermediate activations based on the injected auxiliary information.
    However, their application to camera geometry and distortion handling remains
    unexplored. Different from the distortion-aware architectures, feature modulation
    offers a lightweight and unified approach to adapt features by dynamically conditioning
    intermediate representations on fisheye parameters and distortion maps. This
    enables view-specific adaptation while preserving consistency for feature
    extraction and cross-view fusion in BEV.

    Our work pioneers the integration of FiLM-style conditioning with projection-free
    BEV detection, extending modulation beyond decoder features to include 3D positional
    embeddings of object queries. This creates the first end-to-end, geometry-aware
    solution for multi-camera 3D detection that explicitly addresses fisheye distortion
    through feature modulation.

    \section{METHODOLOGY}
    \label{sec:methodology}

    Our core contribution is a projection-free BEV detection framework, Distortion-Aware
    PETR, designed to handle the geometric challenges of mixed pinhole and
    fisheye camera systems. An overview of our method is shown in Fig.~\ref{fig:full_structure}.
    The framework enhances the PETR architecture with learned distortion-adaptive
    modules, including unified distortion-aware positional embeddings for both
    2D image features and 3D coordinates, and a bidirectional co-modulation mechanism
    that mutually refines image features and positional embeddings, creating a
    unified distortion-aware system while preserving PETR's projection-free
    efficiency.

    \subsection{Distortion Modeling in 3D Position Encoding}
    We adopt the unified MEI camera model~\cite{mei2007single} to handle mixed
    pinhole and fisheye lens distortion in KITTI-360. Unlike traditional pinhole
    cameras that assume rectilinear projection, fisheye cameras exhibit significant
    radial distortion requiring specialized modeling. The MEI model provides a
    mathematically elegant framework unifying pinhole and fisheye cameras
    through a single parameter set, enabling seamless integration across different
    camera types within the same multi-view system~\cite{samani2023f2bev}.

    For a 3D point $(X, Y, Z)$ in camera coordinates, the MEI model first
    projects the point to a unit sphere, then applies perspective projection with
    mirror parameter $\xi$, followed by radial distortion correction and image
    plane projection with $\mathbf{K}_{f}$:
    \begin{equation}
        \begin{aligned}
            \mathbf{P}_{s} & = \mathbf{P}/\lVert\mathbf{P}\rVert                                 \\
            \mathbf{P}_{c} & = \left(\frac{X_{s}}{Z_{s} + \xi}, \frac{Y_{s}}{Z_{s} + \xi}\right) \\
            r^{2}          & = X_{c}^{2}+ Y_{c}^{2}                                              \\
            \mathbf{P}_{d} & = (1 + k_{1}r^{2}+ k_{2}r^{4}) \times \mathbf{P}_{c}                \\
            \mathbf{P}_{I} & = \mathbf{K}_{f}\,\mathbf{P}_{d}
        \end{aligned}
    \end{equation}
    This unified formulation reduces to pinhole projection when $\xi = 0$ and $k_{i}
    = 0$, enabling consistent processing across mixed camera configurations. PETR
    employs 3D position encoding to establish spatial correspondences between image
    features and 3D queries. We replace 3D coordinate generation with a
    distortion-aware ray generation in the position-encoding module to account
    for non-linear distortion characteristics:
    \begin{equation}
        \mathbf{R}_{\text{f}}(u,v,d) = \text{UnprojectMEI}([u,v], \mathbf{K}_{\text{f}}
        , \xi, k_{1}, k_{2}) \times \mathbf{D}_{\text{u}}
    \end{equation}
    where $\text{UnprojectMEI}(\cdot)$ implements the inverse unified camera model
    transformation along uniformly distributed depths $\mathbf{D}_{\text{u}}$.
    Following PETR's 3D Position Encoder~\cite{liu2022petr}, the unprojected 3D coordinates
    are transformed to 3D position embedding by a multi-layer perceptron (MLP) network
    $\psi(.)$. Then the 2D features are flattened and added with position
    embedding to formulate 3D position-aware features for the Transformer
    decoder.
    \begin{equation}
        \mathbf{PE}^{3d}_{i}(t)= \psi\big(\mathbf{R}_{\text{f}}(u,v,d)\big),
    \end{equation}

    \subsection{Feature-Positional Embedding Co-Modulation}
    In the original PETR, the 3D positional embedding is simply added to the
    image features. To create a more powerful fusion of geometric and appearance
    information, we introduce a bidirectional co-modulation (FPECoM) module that
    enables mutual refinement between image features and position embeddings.
    \begin{figure}[htbp]
        \centering
        \includegraphics[width=1.0\linewidth]{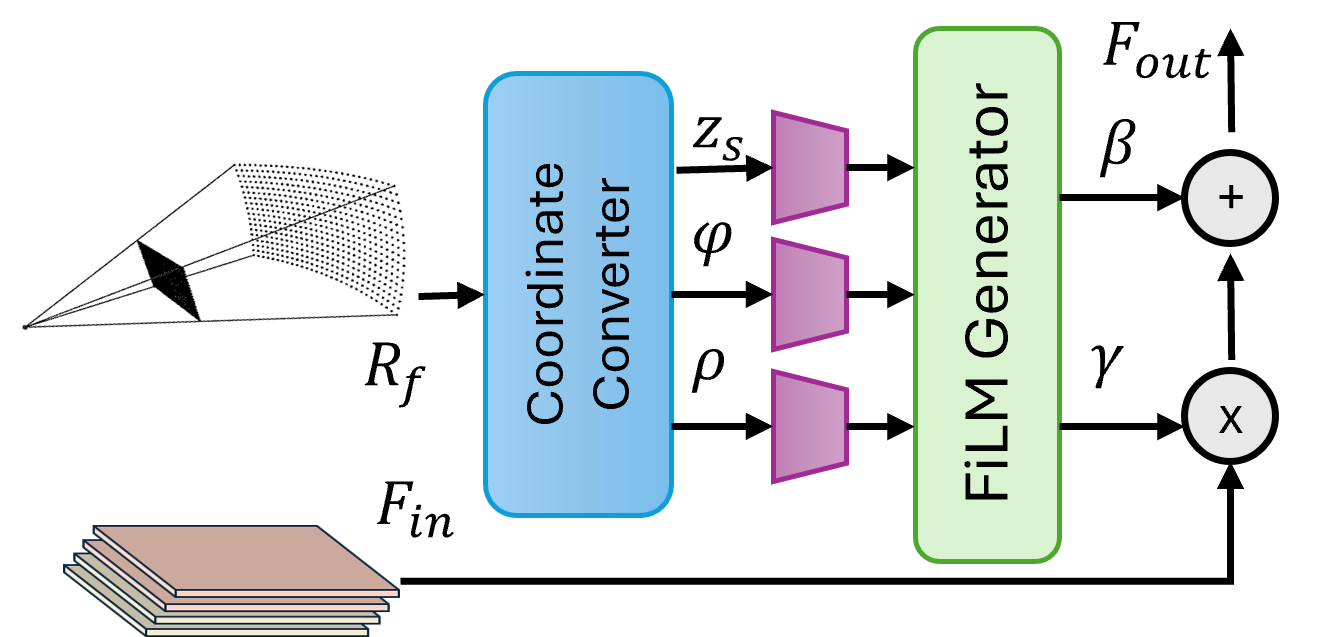}
        \caption{The architecture of our spatial FiLM module. Unprojected rays $(
        x_{s}, y_{s}, z_{s})$ are converted to spherical coordinates (e.g.,
        radial and elevation angles) as inputs. These are processed by separate
        encoders and a FiLM generator network to produce the spatial and
        channel-wise modulation parameters $\beta$ and $\gamma$ as outputs, which
        are then applied to the image features.}
        \label{fig:spatial_film}
    \end{figure}
    Our co-modulation approach operates in two stages using Feature-wise Linear Modulation
    (FiLM)~\cite{perez2018film}. First, we modulate the image features based on the
    distortion map derived from unprojecting image pixels to 3D coordinates $(x_{s}
    , y_{s}, z_{s})$ on a unit sphere. The distortion-derived scaling factor
    $\gamma_{d}$ and bias factor $\beta_{d}$ are computed from the converted
    spherical coordinate in $(\rho,\phi,z_{s})$ format:

    \begin{equation}
        \mathbf{F}_{mod}= \gamma_{d}\odot F_{img}+ \beta_{d},
    \end{equation}
    where $F_{img}$ represents the original image features. Second, the 3D
    positional embedding is further guided by these distortion-modulated image
    features, where feature-aware factors $\gamma_{a}$ and $\beta_{a}$ are derived
    from $F_{mod}$:
    \begin{equation}
        PE_{out}= \gamma_{a}\odot PE_{in}+ \beta_{a},
    \end{equation}
    where $PE_{in}$ represents the distortion-aware 3D positional embedding. This
    bidirectional co-modulation allows the attention mechanism to learn how geometric
    distortion and visual appearance jointly influence spatial reasoning,
    creating a more robust feature-geometry correspondence under varying
    distortion conditions.

    Similarly, PETRv2~\cite{liu2023petrv2} employs feature-guided positional encoding
    (FPE) with SE layers to enhance the positional embeddings to capture input dependency.
    Unlike SE layers, which merely reweight channels through a global gating vector,
    our FiLM-based co-modulation enables spatially adaptive scaling and shifting
    of features and positional embeddings. This richer affine transformation enables
    the network not only to emphasize embedding dimensions, as in FPE, but also to
    reposition them in the latent space based on local distortion and appearance
    cues. By jointly adapting appearance features and geometric priors, the decoder
    achieves better query–feature correspondence. In our experiments, we demonstrate
    that this mutual modulation approach significantly outperforms unidirectional
    alternatives like standard FiLM on features only and SE layers.

    \subsection{Polar 2D Positional Encoding}
    To make the 2D image features aware of the underlying fisheye geometry, we
    introduce a hybrid positional encoding scheme. This encoding augments the standard
    2D positional information with distortion-aware polar coordinates derived from
    the camera's MEI model.

    For each pixel $(u,v)$ in an image from camera $n$, we generate a multi-component
    positional embedding. First, following standard practice in detection
    transformers, we compute sinusoidal embeddings for the pixel's grid position
    $(x, y)$ and the camera index $n$. This provides the network with basic
    spatial and view-identity information.

    Second, to encode the geometric distortion, we unproject each pixel to its
    corresponding 3D ray on the unit sphere, yielding a vector $\mathbf{P}_{s}= (
    x_{s}, y_{s}, z_{s})$. From this, we derive polar coordinates on the sensor plane
    projection:
    \begin{equation}
        r = \sqrt{x_{s}^{2}+ y_{s}^{2}}, \quad \theta = \operatorname{atan2}(y_{s}
        , x_{s})
    \end{equation}
    where $r$ is the radial distance from the principal point on the unit sphere
    projection and $\theta$ is the azimuth angle. These two components explicitly
    describe each pixel's position relative to the lens's optical center,
    directly encoding the non-linear nature of the fisheye projection.

    Following~\cite{carion2020end}, all components are converted into high-frequency
    embeddings using a sinusoidal function
    \begin{equation}
        \begin{aligned}
            \text{PE}(p, 2i) = \sin(p / T^{2i/d}),  \\
            \text{PE}(p, 2i+1) = \cos(p / T^{2i/d})
        \end{aligned}
    \end{equation}
    where $p$ is the input position ($x, y$, $n$, $r, \theta$), $T$ is the
    temperature, $d$ is the feature dimension, and $i$ is the dimension index. By
    combining grid-based coordinates with geometry-derived polar coordinates, we
    provide the cross-attention mechanism with a rich, hybrid representation that
    preserves both the image's raster structure and the camera's intrinsic geometric
    properties, improving feature-query correspondence under severe distortion.

    \begin{table*}
        [!htbp]
        \centering
        \caption{Comprehensive evaluation on full KITTI-360 dataset.}
        \label{tab:comprehensive_results}
        \begin{tabular}{l|c|cccccc|ccc}
            \toprule[2pt] \textbf{Model} & \textbf{Backbone} & \textbf{mAP}$\uparrow$ & \textbf{NDS}$\uparrow$ & \textbf{mATE}$\downarrow$ & \textbf{mASE}$\downarrow$ & \textbf{mAOE}$\downarrow$ & \textbf{mAVE}$\downarrow$ & \textbf{AP}$_{\text{car}}$ $\uparrow$ & \textbf{AP}$_{\text{ped}}$ $\uparrow$ & \textbf{AP}$_{\text{bus}}$ $\uparrow$ \\
            \midrule BEVDet              & ResNet-50         & 0.121                  & 0.159                  & 0.736                     & 0.481                     & 1.031                     & 1.017                     & 0.462                                 & 0.096                                 & 0.000                                 \\
            Polar-BEVDet                 & ResNet-50         & 0.153                  & 0.219                  & 0.655                     & 0.384                     & 0.979                     & \textbf{0.846}            & 0.517                                 & 0.153                                 & 0.002                                 \\
            \midrule BEVFormer           & ResNet-101        & 0.184                  & 0.216                  & 0.716                     & 0.378                     & 0.847                     & 1.216                     & 0.468                                 & 0.131                                 & 0.103                                 \\
            BEVFormer(F2BEV)             & ResNet-101        & 0.167                  & 0.236                  & 0.720                     & 0.370                     & \textbf{0.780}            & 0.982                     & 0.508                                 & 0.136                                 & 0.071                                 \\
            \midrule PETR                & VovNet-99         & 0.272                  & 0.290                  & 0.629                     & 0.340                     & 0.834                     & 0.961                     & 0.578                                 & 0.227                                 & 0.140                                 \\
            PETR(CAMConv)                & VovNet-99         & 0.279                  & 0.299                  & 0.608                     & \textbf{0.339}            & 0.905                     & 0.874                     & 0.602                                 & 0.277                                 & \textbf{0.146}                        \\
            PolarPETR                    & VovNet-99         & 0.280                  & 0.288                  & 0.598                     & 0.347                     & 0.875                     & 0.993                     & 0.602                                 & \textbf{0.296}                        & 0.104                                 \\
            DAPETR                       & VovNet-99         & \textbf{0.286}         & \textbf{0.306}         & \textbf{0.585}            & 0.342                     & 0.816                     & 0.951                     & \textbf{0.603}                        & 0.270                                 & 0.139                                 \\
            \bottomrule[2pt]
        \end{tabular}
    \end{table*}

    \section{EXPERIMENTS}

    Our work leverages the KITTI-360 dataset~\cite{liao2022kitti}, a comprehensive
    dataset featuring both forward-facing pinhole cameras and two $190^{\circ}$ fisheye
    cameras, making it ideal for mixed-camera 3D object detection. To enable standardized
    evaluation with the existing BEV detection frameworks, we utilize the KITTI-360
    to nuScenes conversion pipeline introduced in~\cite{liu2026benchmarking}. We
    refer readers to~\cite{liu2026benchmarking} for comprehensive details regarding
    annotation alignment, calibration adaptation, and evaluation protocol adjustments
    necessary for this conversion.

    We conduct a comprehensive set of experiments on the converted KITTI-360 benchmark
    to validate the effectiveness of DAPETR. Evaluation is performed on 10
    classes: [car, truck, trailer, bus, bicycle, motorcycle, pedestrian, pole,
    object, traffic sign] using nuScenes benchmark metrics without average
    attribute error (AAE) and rebalanced weights for NDS calculation. For the ablation
    studies, we sampled approximately 11k out of 56k frames of 258 scenes for
    agile training. The validation split contains 8500 frames from 41 scenes.

    \subsection{Implementation Details}
    We use a VovNet-99 backbone to extract image features. The transformer
    decoder consists of 6 layers and 900 queries. We train the model for 24 epochs
    using the AdamW optimizer~\cite{loshchilov2017decoupled}. The initial
    learning rate is set to $2 \times 10^{-4}$ and is decayed using a cosine
    annealing schedule. Our model is trained on 2 NVIDIA A5000 GPUs with a batch
    size of 8 and the learning rate is scaled linearly with the batch size. All
    models are implemented within the MMDetection3D~\cite{mmdet3d2020} framework.
    For data preprocessing, we crop and resize the fisheye image to the same
    size as the pinhole image of 1408x376.

    For the other models reported in Table~\ref{tab:comprehensive_results}: Polar-BEVDet
    is adapted from PolarBEVDet without the temporal modeling and auxiliary 2d supervision.
    BEVFormer(F2BEV) is the small static version of BEVFormer without temporal modeling,
    and the VTM is replaced with the fisheye projection from F2BEV. For PETR(CAMConv),
    we append three further channels: $( \rho,\phi,z_{s})$ from the unprojected
    3D rays on the first convolution layer.

    PolarPETR is our re-implementation of PETR in polar coordinates~\cite{liu2026benchmarking}.
    3D PE and object queries of PETR are converted from Cartesian $(x,y,z)$ to cylindrical coordinates.
    For each 3D point generated by PETR's frustum sampling, we compute $(\rho,\theta
    ,z)$ and normalize with the maximum detection range $\rho_{\max}$ and angular
    range $2\pi$, which are fed into the 3D position encoder. Object queries are
    initialized uniformly in polar space and passed directly to the transformer decoder,
    allowing the model to learn polar representations and predict offsets
    relative to polar reference points. For loss computation and the final box regression
    in the detection head, predicted $(\rho,\theta,z )$ and offsets are
    converted back to Cartesian coordinates to maintain compatibility with
    standard 3D detection annotations and evaluation protocols.

    \subsection{Main Results}
    We benchmark several state-of-the-art BEV detection frameworks on our
    converted KITTI-360 dataset, with results summarized in Table~\ref{tab:comprehensive_results}.

    First, we evaluate representative projection-based models. Both forward-projection
    (BEVDet) and backward-projection (BEVFormer) methods struggle with the native
    fisheye imagery, achieving only 0.121 and 0.184 mAP, respectively. While
    their distortion-aware (F2BEV) or polar variants (Polar-BEVDet) bring modest
    gains, they remain significantly outperformed by projection-free approaches,
    validating our choice to build upon PETR.

    Our main analysis focuses on PETR variants. The standard PETR baseline on native
    fisheye images achieves a strong 0.272 mAP. We evaluate existing camera-aware
    adaptations like CAM-Conv, which provides a slight improvement. The explicit
    geometric approach, PolarPETR, also boosts performance, confirming the benefits
    of a polar representation.

    Finally, our proposed model, DAPETR, which integrates our learned distortion-aware
    modules, achieves a new state-of-the-art on the native mixed-camera setup.
    It reaches 0.286 mAP and 0.306 NDS, demonstrating performance surpassing baseline.
    The performance is significant, as our method handles severe fisheye distortion
    natively without rectification, preserving the full field of view and image
    information.

    \textbf{Per-class AP analysis} reveals a large performance disparity across classes
    in KITTI-360 dataset. While all models perform well on common classes like cars,
    they struggle on less frequent classes like buses, due to the extremely limited
    samples (cars: 430K, buses: 1K). Although the number of pedestrian samples is
    comparable to nuScenes dataset, the performance gap is still large, which
    reveals the challenge of detecting small objects with fisheye images.

    \textbf{Runtime and Efficiency Analysis:} To validate our claims of efficiency,
    we compare the computational costs of DAPETR against the baseline models in
    Table~\ref{tab:efficiency}. Inference speeds (FPS) and
    detection head GFLOPs are measured on a single GPU with a
    batch size of 1. As shown, DAPETR introduces nominal overhead: bidirectional feature modulation adds merely 0.57M parameters and 4.67 GFLOPs to the detection
    head compared to the baseline. Meanwhile, it maintains a highly
    competitive inference speed of 6.8 FPS (compared to 7.0 FPS for PETR). This demonstrates
    that our distortion-aware mechanism preserves the inherent efficiency of projection-free
    architectures while significantly boosting detection performance.

    \begin{table}[h]
        \centering
        \caption{Computational Cost Comparison}
        \label{tab:efficiency}
        \begin{tabular}{lccc}
            \toprule Method & Params (M) & Head GFLOPs & FPS \\
            \midrule PETR   & 81.80      & 51.27       & 7.0 \\
            PolarPETR       & 82.06      & 53.39       & 7.7 \\
            DAPETR (Ours)   & 82.37      & 55.94       & 6.8 \\
            \bottomrule
        \end{tabular}
    \end{table}

    \subsection{Ablation Studies}
    We perform a series of ablation studies to analyze the contribution of each
    component, with results shown in Table~\ref{table:ablation_polar}. Our
    baseline is PETR applied to native fisheye images, where only the fundamental
    distortion model (DM) for ray unprojection is used. This performs worse than
    the baseline on rectified images (23.3 vs 25.0 mAP), highlighting the limit of
    only distortion modeling. With further enhancements, we all observed consistent
    transcendence over the rectified baseline on the small-scale dataset.

    \textbf{Effect of Polar Representation.} Introducing a polar representation
    for BEV queries and 3D coordinates improves performance significantly over
    the baseline (+1.6 mAP). This confirms that explicitly aligning the query
    space with the radial distortion nature of fisheye geometry provides a good
    inductive bias.

    \textbf{Effect of Co-Modulation.} Independently, spatial modulation (SM)
    and feature-guided modulation (FgM) also improve performance. For FgM, using
    FiLM is notably more effective than a simpler SE layer (+0.8 mAP),
    demonstrating the benefit of its affine transformation for fusing geometric
    and appearance information.

    \begin{table}[h]
        \caption{Ablation Study on Model Components. DM: distortion modelling, PR:
        polar representation, SM: spatial modulation, FgM: feature-guided
        modulation, PPE: polar positional encoding. The first row indicates the performance
        of baseline on rectified images of KITTI-360.}
        \label{table:ablation_polar}
        \centering
        \begin{tabular}{c|c|c|c|c|cc}
            \toprule DM         & PR         & SM         & FgM  & PPE        & mAP$\uparrow$ & NDS$\uparrow$ \\
            \midrule            &            &            &      &            & 25.0          & 25.3          \\
            \midrule \checkmark &            &            &      &            & 23.3          & 25.7          \\
            \checkmark          & \checkmark &            &      &            & 24.9          & 26.8          \\
            \checkmark          &            & \checkmark &      &            & 25.2          & 26.8          \\
            \checkmark          &            & \checkmark & SE   &            & 25.4          & 26.0          \\
            \checkmark          &            & \checkmark & FiLM &            & 26.2          & 27.8          \\
            \checkmark          &            & \checkmark & FiLM & \checkmark & \textbf{26.3} & \textbf{28.5} \\
            \checkmark          & \checkmark & \checkmark & FiLM & \checkmark & 25.5          & 27.5          \\
            \bottomrule
        \end{tabular}
    \end{table}

    \textbf{Spatial FiLM Position Ablation:} We further study where to place the
    spatial FiLM module that uses a distortion map to modulate image features
    for distortion awareness. Intuitively, conditioning early in the network
    should allow the model to normalize distortion from the start and propagate
    geometry-aware features through all subsequent stages. However, our results
    show the opposite trend: applying spatial FiLM at the detection head yields the
    best performance, while inserting it in the backbone or neck underperforms (Table~\ref{tab:film_position}).
    We hypothesize that early modulation may overfit low-level textures to camera-specific
    distortion and harm feature generality across views; the neck mixes multi-scale
    features and may dilute the signal. In contrast, head-level modulation acts
    closest to the cross-attention and box decoding, aligning geometry and appearance
    precisely where queries meet features. This also keeps the shared backbone
    features largely camera-agnostic, improving cross-camera consistency.

    \begin{table}[h]
        \caption{Spatial FiLM placement ablation on KITTI-360. Head-level
        placement performs best, contrary to the hypothesis that "earlier is
        better."}
        \label{tab:film_position}
        \centering
        \begin{tabular}{l|cc}
            \toprule Placement & mAP $\uparrow$ & NDS $\uparrow$ \\
            \midrule Backbone  & 23.5           & 25.7           \\
            Neck               & 23.3           & 25.9           \\
            Head               & \textbf{25.4}  & \textbf{26.0}  \\
            \bottomrule
        \end{tabular}
    \end{table}
    \textbf{Effect of Polar Positional Encoding.} Building on the best co-modulation
    model, adding the polar positional encoding (PPE) for 2D image tokens further
    improves performance, achieving our best result of 26.3 mAP and 28.5 NDS. This
    shows that providing the 2D features with explicit, token-wise information
    about the fisheye geometry is crucial for robust cross-attention.

    \textbf{Conflicting Interaction with Polar Representation.} However, when we
    combine the explicit polar representation (PR) with our best-performing
    model (SM + FgM + PPE), performance unexpectedly decreases by 0.8 mAP. This
    suggests a conflict between the two strategies. The learned co-modulation and
    positional encoding likely create their own implicit correction for the fisheye
    geometry, which becomes redundant or even counter-productive when applied to
    a BEV space that is already explicitly warped into polar coordinates. This finding
    indicates that explicit geometric re-parameterization and learned feature adaptation
    should be treated as competing, rather than complementary, design choices.
    \subsection{Range and Angular Stratified Results}
    We evaluate the detection performance tiered by distance and angle. As shown
    in Table~\ref{tab:range_map}, both PolarPETR and our DAPETR outperform the
    baseline PETR across all distances. Notably, DAPETR shows the most
    significant gains in the 10-30m range, improving mAP by over 1.4\% compared
    to PolarPETR and 5.5\% compared to the baseline PETR. This suggests that our
    learned distortion-aware modules are particularly effective at handling the
    smaller object sizes at mid-range distances.

    Similarly, the angular-stratified results in Table~\ref{tab:angular_map} demonstrate
    the effectiveness of our approach in full $360^{\circ}$ perception. DAPETR
    achieves the highest mAP in the front and back sectors, with particularly notable
    gains in the front sector (+2.35\% over baseline PETR). In the back sector, which
    is primarily covered by fisheye cameras, DAPETR improves mAP by 1.99\% over the
    baseline, confirming that our learned distortion-aware modules effectively
    mitigate fisheye distortions. For the side views, PolarPETR achieves the
    best performance at 29.84\%, outperforming DAPETR by 1.17\%. This suggests
    that explicit polar representation may offer advantages for certain viewing
    angles, while our learned approach excels in the more challenging front and back
    regions where accurate depth estimation and feature-geometry alignment are critical
    for detection accuracy.

    \begin{table}[t]
        \centering
        \caption{mAP (\%) across distance ranges on KITTI-360 mixed pinhole-fisheye
        images.}
        \label{tab:range_map} \resizebox{\linewidth}{!}{
        \begin{tabular}{lccccc}
            \toprule Method & 0--10\,m & 10--20\,m & 20--30\,m & 30--40\,m & 40--50\,m \\
            \midrule PETR   & 54.42    & 30.75     & 12.01     & 4.21      & 1.01      \\
            PolarPETR       & 56.57    & 35.52     & 12.49     & 4.50      & 1.44      \\
            DAPETR          & 57.29    & 36.27     & 13.94     & 4.85      & 1.36      \\
            \bottomrule
        \end{tabular}
        }
    \end{table}

    \begin{table}[t]
        \centering
        \caption{Angular-stratified mAP (\%) on KITTI-360 mixed pinhole-fisheye
        images.}
        \label{tab:angular_map} \resizebox{\linewidth}{!}{
        \begin{tabular}{lccc}
            \toprule Method & Front $120^{\circ}$ & Back $120^{\circ}$ & Sides $120^{\circ}$ \\
            \midrule PETR   & 29.84               & 21.87              & 28.77               \\
            PolarPETR       & 31.35               & 22.68              & 29.84               \\
            DAPETR          & 32.19               & 23.86              & 28.67               \\
            \bottomrule
        \end{tabular}
        }
    \end{table}
    \begin{figure}[htbp]
        \centering
        \includegraphics[width=1.0\linewidth]{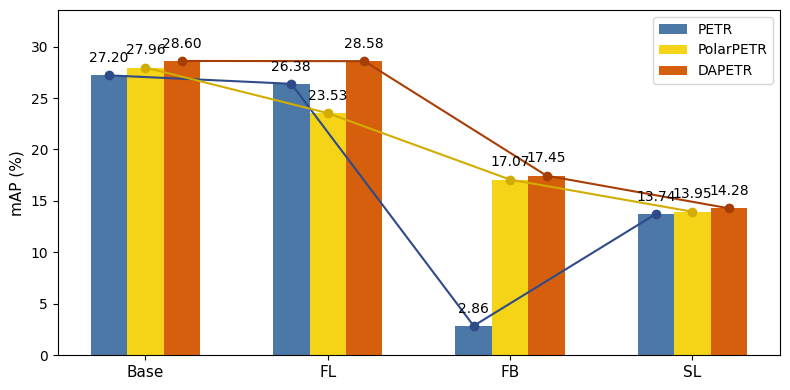}
        \caption{Model robustness under camera failure scenarios. FL, FB and SL
        denote the front-left, front-both and side-left.}
        \label{fig:camera_robustness}
    \end{figure}
    \subsection{Camera Loss Robustness}
    To evaluate the fault tolerance of our approach, we test model robustness under
    camera failure scenarios by randomly mocking cameras during inference. As
    shown in Fig.~\ref{fig:camera_robustness}, for front-left camera failure (FL),
    all methods show relatively modest degradation, with DAPETR maintaining
    28.58\% mAP compared to 26.38\% for PETR. This suggests that both front cameras
    can partially compensate for the missing view. Although the FOVs of both
    front cameras overlap significantly, the BEV object detection does not benefit
    much from the redundancy.

    The most significant degradation occurs when both front cameras are lost (FB),
    leaving only the side fisheye cameras for perception. Here, DAPETR
    demonstrates superior resilience with 17.45\% mAP, substantially
    outperforming both PETR (3.86\%). Without specifically designed distortion-aware
    modules, the baseline PETR nearly fails completely, as it cannot effectively
    leverage the fisheye views alone.

    For side-left camera failure (SL), all methods show graceful degradation,
    with DAPETR still achieving the highest mAP. The relatively smaller performance
    drop compared to front camera failures indicates that the remaining cameras can
    provide sufficient coverage for reasonable detection performance. Overall, these
    results confirm that DAPETR's learned distortion adaptation creates more robust
    feature representations when parts of the sensor suite fail.

    \subsection{Qualitative Results}
    \begin{figure}[h]
        \centering
        \includegraphics[width=\linewidth]{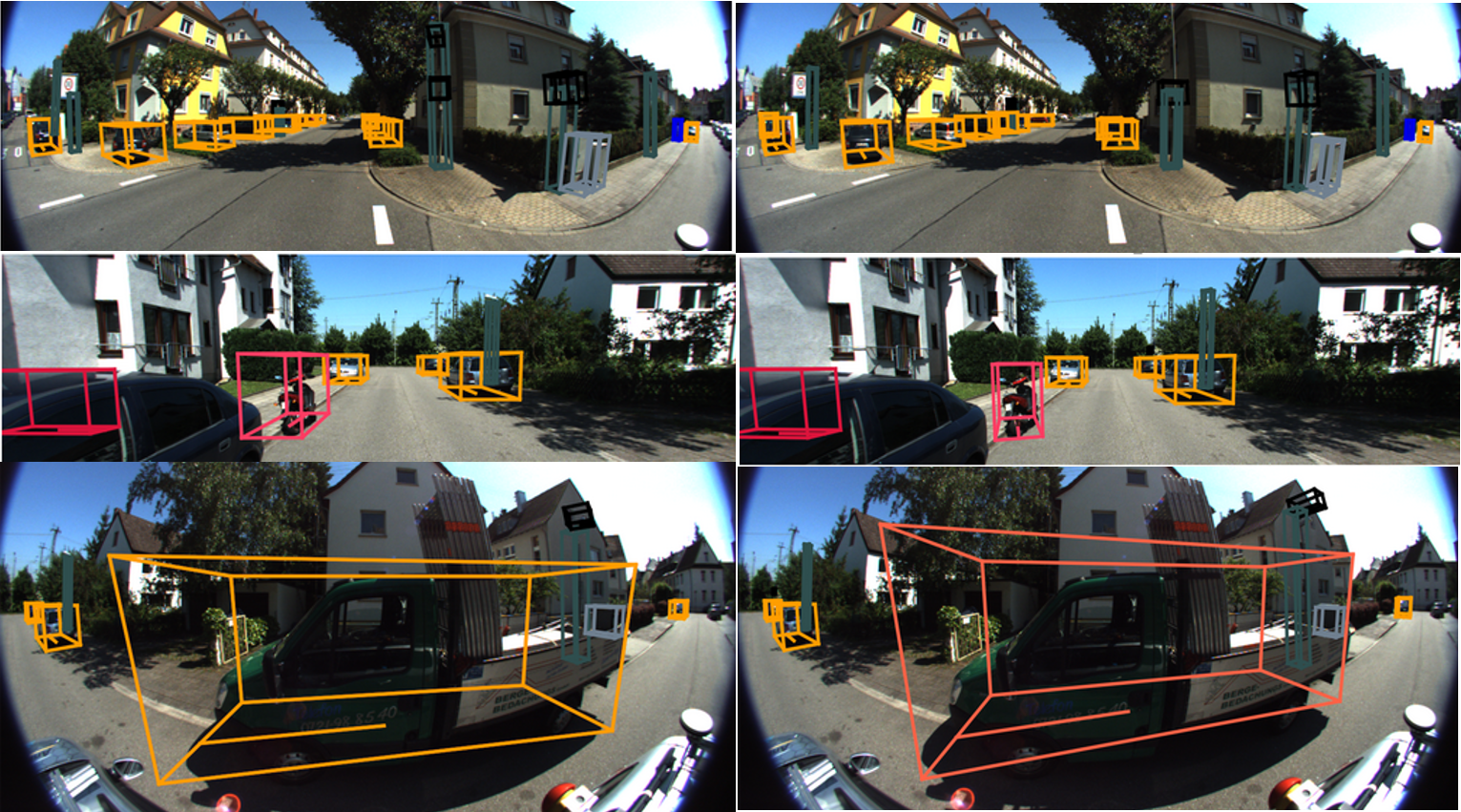}
        \caption{Qualitative comparison between the baseline PETR (left) and our
        DAPETR (right). Our model demonstrates improved localization and classification
        accuracy. From top to bottom: better rotation estimation for cars,
        tighter scale for a motorcycle, and the correct classification of a truck,
        which the baseline misses.}
        \label{fig:qualitative_results}
    \end{figure}

    As illustrated in Fig.~\ref{fig:qualitative_results}, our DAPETR model
    produces significantly improved qualitative results compared to the baseline
    PETR. The left column shows the baseline's predictions, while the right
    column shows our model's predictions. In the top row, DAPETR demonstrates
    more accurate orientation estimation for cars in the near distance. The middle
    row highlights better scale prediction, with our model producing a much
    tighter bounding box for the motorcycle. Finally, the bottom row shows a
    clear case of improved classification and detection; our model correctly identifies
    the large vehicle as a truck, whereas the baseline fails to detect it, showcasing
    the benefits of our distortion-aware approach in challenging fisheye views.

    \section{CONCLUSION}

    In this paper, we addressed the challenge of 3D object detection in mixed pinhole-fisheye
    camera systems by proposing Distortion-Aware PETR (DAPETR). Instead of
    relying on explicit geometric transformations or rectification, DAPETR incorporates
    two novel learned-adaptive modules: a unified distortion-aware positional
    embedding using MEI for both 2D image features and 3D position coordinates, and
    a bidirectional co-modulation module to mutually refine image features and 3D
    positional embeddings. Through extensive experiments on the converted KITTI-360
    dataset, we found that projection-free methods (PETR) prove most adaptable,
    achieving the highest mAP compared to projection-based methods (variants of BEVDet
    and BEVFormer). Our robustness analysis confirms that DAPETR maintains superior
    performance even under camera failure scenarios, retaining viability with
    fisheye-only perception.

    Although polar re-parameterization has been a common strategy for BEV object
    detection, even we proved its effectiveness in fisheye settings, we
    discovered a negative interaction when combining it with our feature
    modulation approach. This suggests that learned feature adaptation and explicit
    polar transformation are competing, rather than complementary choices. Our
    DAPETR model, relying on learned adaptation, achieves the best performance in
    all our experiments, particularly in challenging
    mid-range distances and fisheye-covered views. This work not only
    delivers a fisheye 3D object detector without rectification but also
    provides practical guidance for future research in distortion-aware
    perception. Future work will focus on addressing the remaining challenges in
    detecting small objects and assessing cross-dataset generalizability on other fisheye datasets.


    \bibliographystyle{ieeetr}
    \bibliography{IEEEfull}
\end{document}